\documentclass[letterpaper, 10 pt, conference]{ieeeconf}
\IEEEoverridecommandlockouts
\usepackage[top=54pt,bottom=54pt,left=54pt,right=54pt]{geometry}
\usepackage{amsmath,amssymb,amsfonts}
\usepackage{algorithmic}
\usepackage{graphicx}
\usepackage{textcomp}
\usepackage{xcolor}
\usepackage{tabularx}
\usepackage{multirow}
\usepackage{multicol}
\usepackage{booktabs} 
\usepackage{hyperref}
\usepackage{soul}
\usepackage{bm}
\usepackage{cite}
\hypersetup{
    colorlinks=true,
    linkcolor=blue,
    filecolor=magenta,      
    urlcolor=cyan
}

\usepackage{subcaption}

\newcommand{\tb}{\textbf}

\begin{document}

\title{Quadruped Agile Locomotion:\\ Sparsely Gated Mixture of Experts with Visual Input 
}


\author{Michael Ziegltrum$^{1}$, Jianhao Jiao$^{1}$, Tianhu Peng$^{1}$, Chengxu Zhou$^{1}$, Dimitrios Kanoulas$^{1}$
\thanks{$^{1}$Robot Perception and Learning Lab, Intelligent Robotics, Department of Computer Science, University College London, Gower Street, WC1E 6BT, London, UK. {\tt\small \{michael.ziegltrum.24, jianhao.jiao, tianhu.peng.24, chengxu.zhou, d.kanoulas\}@ucl.ac.uk}}.%
\thanks{This work was supported by the UKRI Future Leaders Fellowship [MR/V025333/1] (RoboHike). For the purpose of Open Access, the author has applied a CC BY public copyright license to any Author Accepted Manuscript version arising from this submission.}
}

\maketitle

\begin{abstract}
Agile locomotion is important for quadruped robots in the real world. Recent approaches have demonstrated impressive capabilities including dynamic climbing and jumping but typically rely on sequential multilayer perceptron (MLP) architectures with densely activated layers. In contrast, sparsely gated mixture-of-experts (MoE) architectures have become popular in the large language model (LLM) domain as a tool to improving scalability and performance by activating only a subset of parameters at inference. In this work, we investigate the application of such architectures to vision-based robotic locomotion. We compare policies based on standard MLPs and MoE architectures under a controlled setting where the number of active parameters at inference time is matched. Experimental results show an 8\% increase in reward during training comparing the MoE to a standard MLP and even show benefits comparing to MLPs with equivalent total parameters. Experimental results on a real Unitree Go2 quadruped robot demonstrate performance in the real world. These results highlight that sparsely gated MoE architectures may pose advantages in tasks where network capacity is important and that borrowing techniques from the fast-moving LLM space and applying them to locomotion is a promising direction.
\end{abstract}



\section{Introduction}
Legged robots pose enormous potential to help humanity in contexts such as search-and-rescue or planetary exploration. These applications require locomotion on challenging terrain. In contrast to walking on flat terrain, obstacles pose significant challenges due to highly dynamic motion and interaction with the environment. It is therefore a pertinent problem to study in order to advance legged robotic utility to humans.

\begin{figure}[t]
  \centering
  \begin{subfigure}{\columnwidth}
    \centering
    \includegraphics[width=\linewidth]{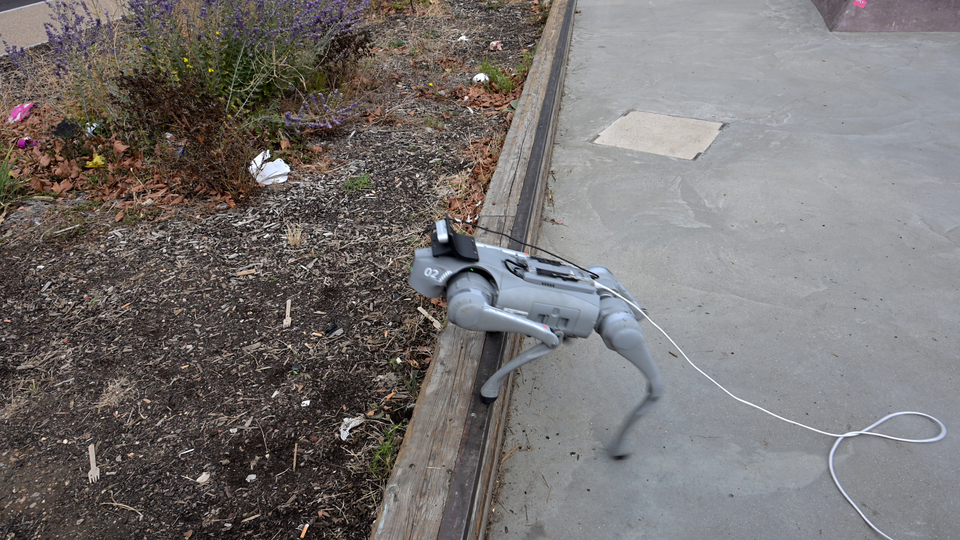}
  \end{subfigure}%
  \caption{The mixture of experts works outside of laboratory conditions in environments with noisier lighting and obstacles that are dissimilar to the training obstacles including cluttered environments.}
  \label{fig:parkour_demos}
\end{figure}

Current approaches to agile locomotion demonstrate impressive results on a variety of terrain, even showing capabilities like bipedal locomotion or combining locomotion with advanced manipulation such as badminton~\cite{hoeller2024anymal,cheng2024extreme,zhuang2023robot,ma2025learning,luo2024pieparkourimplicitexplicitlearning}. Reinforcement learning (RL) and actor-critic methods are commonly used in these examples, but the dominant architecture choice for the actor is a sequential multilayer perceptron (MLP). In the large language model space recent advances have explored other architectures like the sparsely-gated mixture of experts (MoE) and have shown the benefits in scaling and performance~\cite{vaswani2017attention,lepikhin2020gshard,jiang2024mixtral}. In locomotion, while mixing experts has been explored, typical implementations leverage pre-trained experts where experts specialize per skill and are selected at inference time \cite{hoeller2024anymal} or are distilled into one actor policy \cite{caluwaerts2023barkourbenchmarkinganimallevelagility}. In our framework the specialization happens in latent space as a layer in the actor network, differentiating it. Additionally unlike existing MoE work \cite{huang2025moe} we leverage sparsity of experts and vision which to the best of our knowledge have not been incorporated. Our goal is to take advantage of conditional compute via sparsity in the vision-based quadruped locomotion space by training a sparsely-gated MoE. In this network the number of total parameters is greater than the number of parameters activated at each inference, enabling efficient scaling. Our work is most closely related to Extreme Parkour~\cite{cheng2024extreme}, but we modify it in several key ways including most importantly the MoE Actor, noising observations, a depth image pipeline, and other modifications. The main contributions of this paper include: 

\begin{itemize}
  \item A novel framework for leveraging a sparsely-gated mixture of experts in an agile-locomotion context.
  \item Incorporation of perception via a depth camera into the framework.
  \item A robust deployment on the Unitree Go2 Quadruped that includes examples of real world performance on challenging terrain like that of Fig.~\ref{fig:parkour_demos}.
\end{itemize}

We conduct evaluations in training time metrics and in the real world on a challenging obstacle that is 80\% of the robot's height and test various architectures. We find benefits to the MoE by comparing it with a sequential MLP with the same number of parameters as the MoE at inference time, in which the MoE achieved 8\% more reward during training. The MLP with equivalent total parameters is closer in reward to the MoE. Additionally, locomotion is a cyclical task and we analyze expert weighting to find evidence of cyclical expert specialization indicating the possibility of experts specializing in certain parts of the locomotion cycle. The results show that our proposed architecture offers benefits in the vision-based quadruped locomotion space and this and other techniques from the fast-moving LLM space merit further consideration.

The rest of this paper is organized as follows: Sec.~\ref{Sec:RW} presents a review of related work in traditional methods such as model-based control and elevation maps, model-free methods such as RL, and mixture of experts; Sec.~\ref{Sec:Method} is the introduction of the methodology with attention to the MoE architecture; Sec.\ref{Sec:Results} is the evaluation of the method on a real robot and comparison to baselines; and finally Sec.\ref{Sec:Concl} is the conclusion with notes on drawbacks and future work.

\section{Related Works}\label{Sec:RW}

\begin{figure*}[t]
    \begin{subfigure}{0.24\textwidth}
        \begin{subfigure}{1.0\textwidth}
        \includegraphics[width=1.0\textwidth]{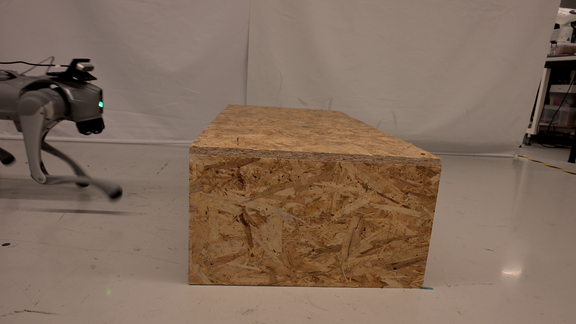}
        \caption{}
        \label{fig:composite_approach}
        \end{subfigure}
    
        \begin{subfigure}{1.0\textwidth}
            \begin{subfigure}{0.45\textwidth}
            \includegraphics[width=1.0\textwidth]{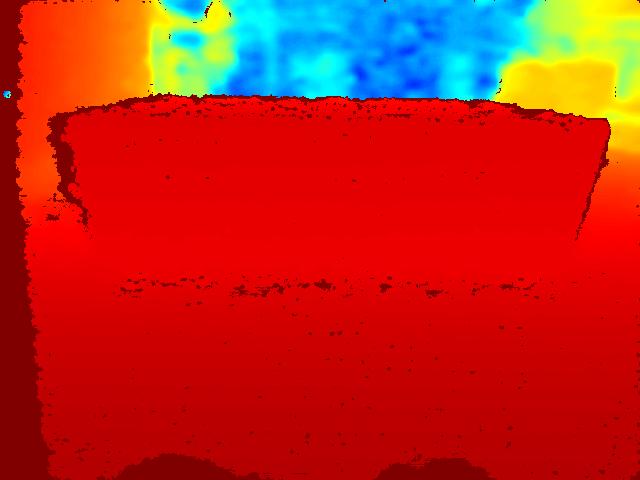}
            \end{subfigure}
            \hfill
            \begin{subfigure}{0.45\textwidth}
            \includegraphics[width=1.0\textwidth]{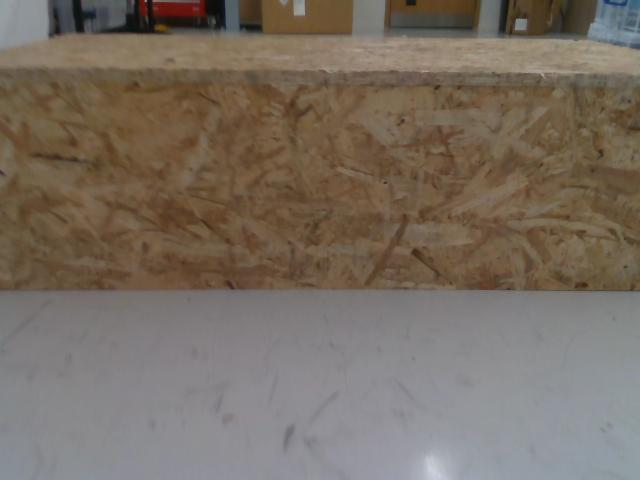}
            \end{subfigure}
        \end{subfigure}
    \end{subfigure}
    \begin{subfigure}{0.24\textwidth}
        \begin{subfigure}{1.0\textwidth}
        \includegraphics[width=1.0\textwidth]{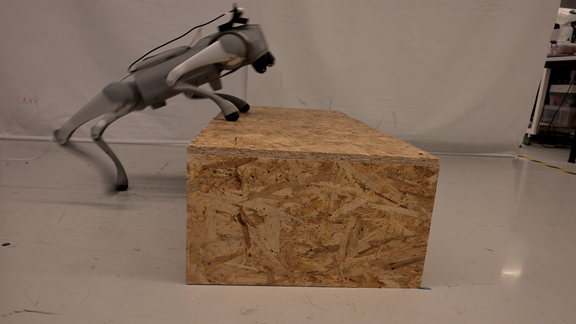}
        \caption{}
        \label{fig:composite_up}
        \end{subfigure}
    
        \begin{subfigure}{1.0\textwidth}
            \begin{subfigure}{0.45\textwidth}
            \includegraphics[width=1.0\textwidth]{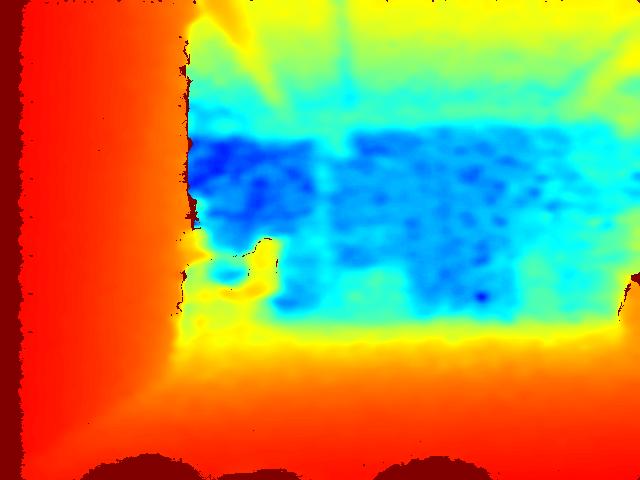}
            \end{subfigure}
            \hfill
            \begin{subfigure}{0.45\textwidth}
            \includegraphics[width=1.0\textwidth]{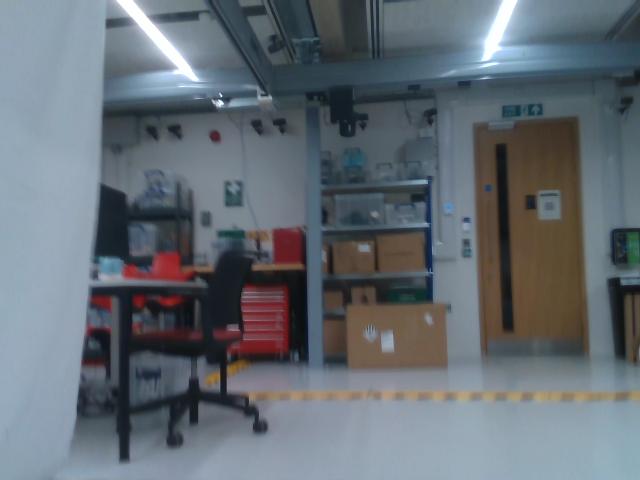}
            \end{subfigure}
        \end{subfigure}
    \end{subfigure}
    \begin{subfigure}{0.24\textwidth}
        \begin{subfigure}{1.0\textwidth}
        \includegraphics[width=1.0\textwidth]{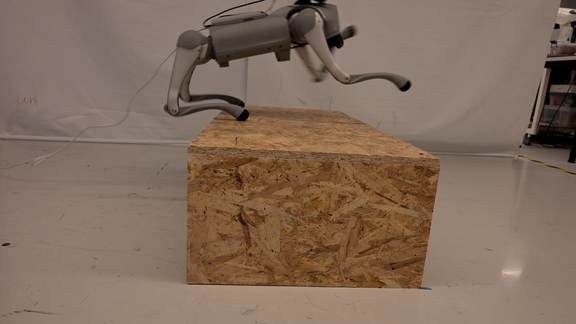}
        \caption{}
        \label{fig:composite_on}
        \end{subfigure}
    
        \begin{subfigure}{1.0\textwidth}
            \begin{subfigure}{0.45\textwidth}
            \includegraphics[width=1.0\textwidth]{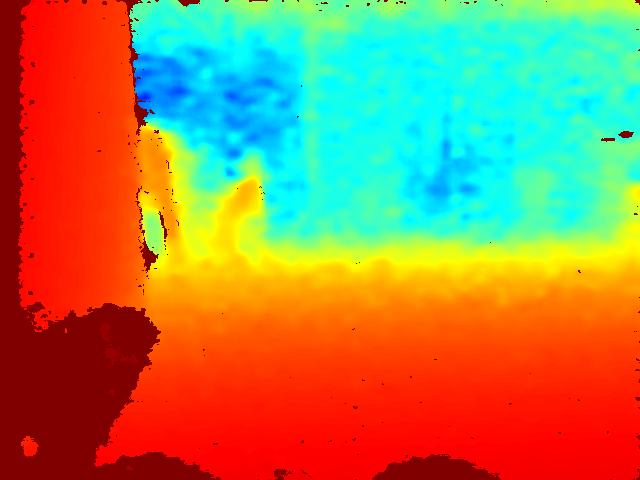}
            \end{subfigure}
            \hfill
            \begin{subfigure}{0.45\textwidth}
            \includegraphics[width=1.0\textwidth]{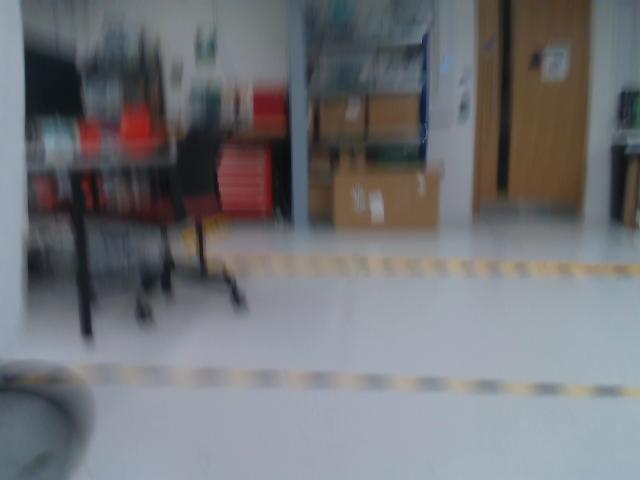}
            \end{subfigure}
        \end{subfigure}
    \end{subfigure}
    \begin{subfigure}{0.24\textwidth}
        \begin{subfigure}{1.0\textwidth}
        \includegraphics[width=1.0\textwidth]{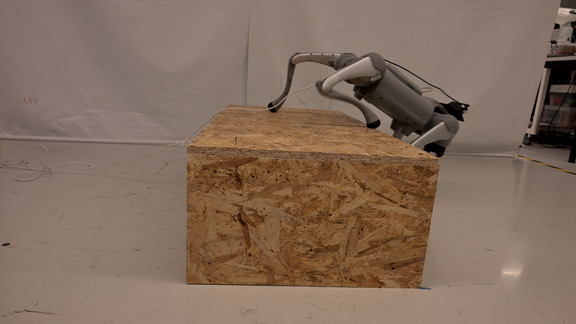}
        \caption{}
        \label{fig:composite_off}
        \end{subfigure}
    
        \begin{subfigure}{1.0\textwidth}
            \begin{subfigure}{0.45\textwidth}
            \includegraphics[width=1.0\textwidth]{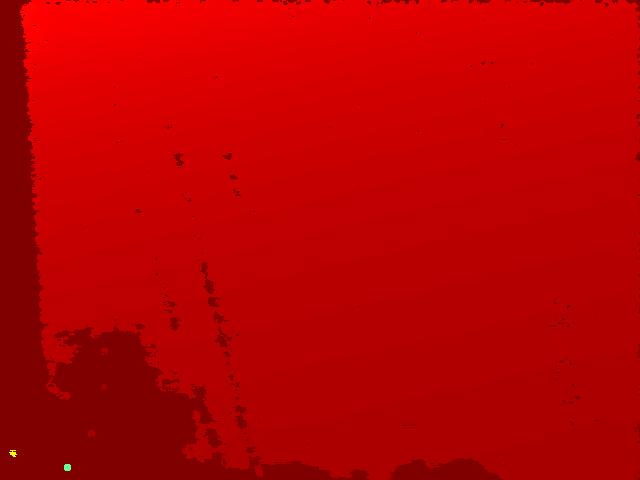}
            \end{subfigure}
            \hfill
            \begin{subfigure}{0.45\textwidth}
            \includegraphics[width=1.0\textwidth]{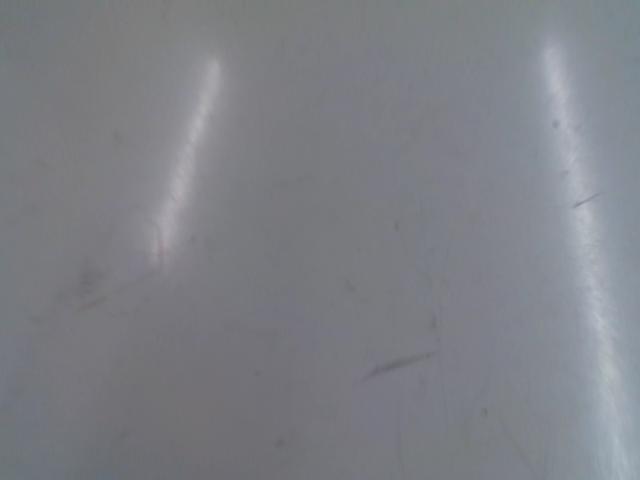}
            \end{subfigure}
        \end{subfigure}
    \end{subfigure}
  \caption{The 32cm Box is shown with the mixture of experts policy controlling the robot. Four key moments are identified where the robot approaches the box \ref{fig:composite_approach}, gets its front legs on the box \ref{fig:composite_up}, stands on the box \ref{fig:composite_on}, and dismounts \ref{fig:composite_off}. Depth images are shown without any processing.}
  \label{fig:parkour_composite}
\end{figure*}

\subsection{Traditional Methods}
Traditional approaches to vision-based agile locomotion often decouple locomotion and perception. The locomotion module may leverage model-based techniques from early examples like Raibert's Hopping Machine~\cite{raibert1984experiments} that treated the system like an inverted pendulum to more recent work like Bledt et al.~\cite{bledt2018cheetah} using a floating base model to climb stairs on the Cheetah Robot~\cite{bledt2018cheetah}, Hutter's et al.~\cite{hutter2016anymal} on Anymal leveraging a floating base model to climb tall stairs, Katz et al.~\cite{katz2019mini} and a floating base model on the Mini Cheetah to perform backflips, Kim et al.~\cite{kim2019highly} blending a floating base model and full-body dynamics to sprint, Bjelonic et al.~\cite{bjelonic2020rolling} applying floating base and whole-body control to work in the DARPA Subterranean Challenge, and Corbères et al.~\cite{10949075} and their work leveraging only whole-body models. For perception typically a separate module is implemented leveraging elevation maps like early examples of Kweon et al.~\cite{kweon1989terrain} prototyping a mars explorer by working in spherical-polar
space, Fankhauser et al.~\cite{fankhauser2018robust} using terrain maps to get footholds and then that as an input into the locomotion controller on rough terrain, and Kim et al.~\cite{9196777} leveraging depth camera pointclouds to construct an elevation map for obstacle avoidance and footstep planning as an input into the locomotion controller. Traditional methods are often tailored to a particular skill, making it difficult to adapt to new skills. They also require complex modeling, which may make it difficult to generalize to new terrain that was not accounted for. In contrast, our method using a model-free approach can accommodate multiple skills and terrain.

\subsection{Learning-based Methods}
Unlike traditional methods, learning-based approaches to agile vision-based locomotion are often model-free and may separate perception and locomotion, or in recent years more likely couple them in an end-to-end approach. Rudin et al.~\cite{rudin2022advancedskillslearninglocomotion} use sampled points from a terrain map as an input into a RL controller to climb and step, Hoeller et al.~\cite{hoeller2024anymal} leverage hierarchical RL learning a module to output maps and inputting that into a learned controller to climb and jump, Agarwal et al.~\cite{agarwal2023legged} use an egocentric camera and end-to-end learning with a depth camera to traverse a variety of terrains, Rudin et al.~\cite{rudin2025parkour} end-to-end learning and multi-expert distillation with RL fine tuning to enable locomotion on challenging terrains, Zhuang et al.~\cite{zhuang2023robot} use end-to-end learning and soft dynamics constraints to jump and run, Cheng et al.~\cite{cheng2024extreme} use end-to-end learning and a unified reward function to enable leaping and climbing. These approaches work and show impressive results on a variety of terrains, but with the exception of recurrent neural networks often used to encode depth images, the dominant approach is a sequential multilayer perceptron. Attention to network size and scaling potential has been limited and our method focuses on solving this using techniques from the large language model space, such as a mixture of experts.


\subsection{Mixture of Experts}
Mixture of experts are popular in the large language model (LLM) space and early examples like that of Jacobs et al.~\cite{6797059} leveraged several expert networks and a gating network to solve vowel discrimination tasks, while more current examples built on this including work by Shazeer et al.~\cite{shazeer2017outrageously} on sparsely gated MoEs that leverage conditional computation on a subset of experts to achieve performance benefits in machine translation. Conditional computation has been widely applied in LLMs like work of Lepikhin et al. with Gshard~\cite{lepikhin2020gshard} to speed up parallel computation and show improvements in machine translation, Zhu et al.~\cite{zhu2024llama} with Llama-moe taking existing dense LLMs and re-training to an MoE to outperform, Du et al.~\cite{du2022glam} with GLAM lowering training costs with an MoE, and the work of Jiang et al.~\cite{jiang2024mixtral} with the Mixtral-8x7B model leveraging conditional compute to outperform on mathematics and other tasks. Until now, these methods have not been widely applied to locomotion with the distinct exception of Huang et al.~\cite{huang2025moe} who apply MoE without vision or conditional computation. Our method relies on conditional compute via a sparsely gated MoE to show performance benefits for vision-based parkour, and unlike \cite{huang2025moe} we apply this as a layer in the latent space of the actor network.

\section{Methods}\label{Sec:Method}

\begin{figure*}[t]
    \centering
    \includegraphics[width=0.98\linewidth]{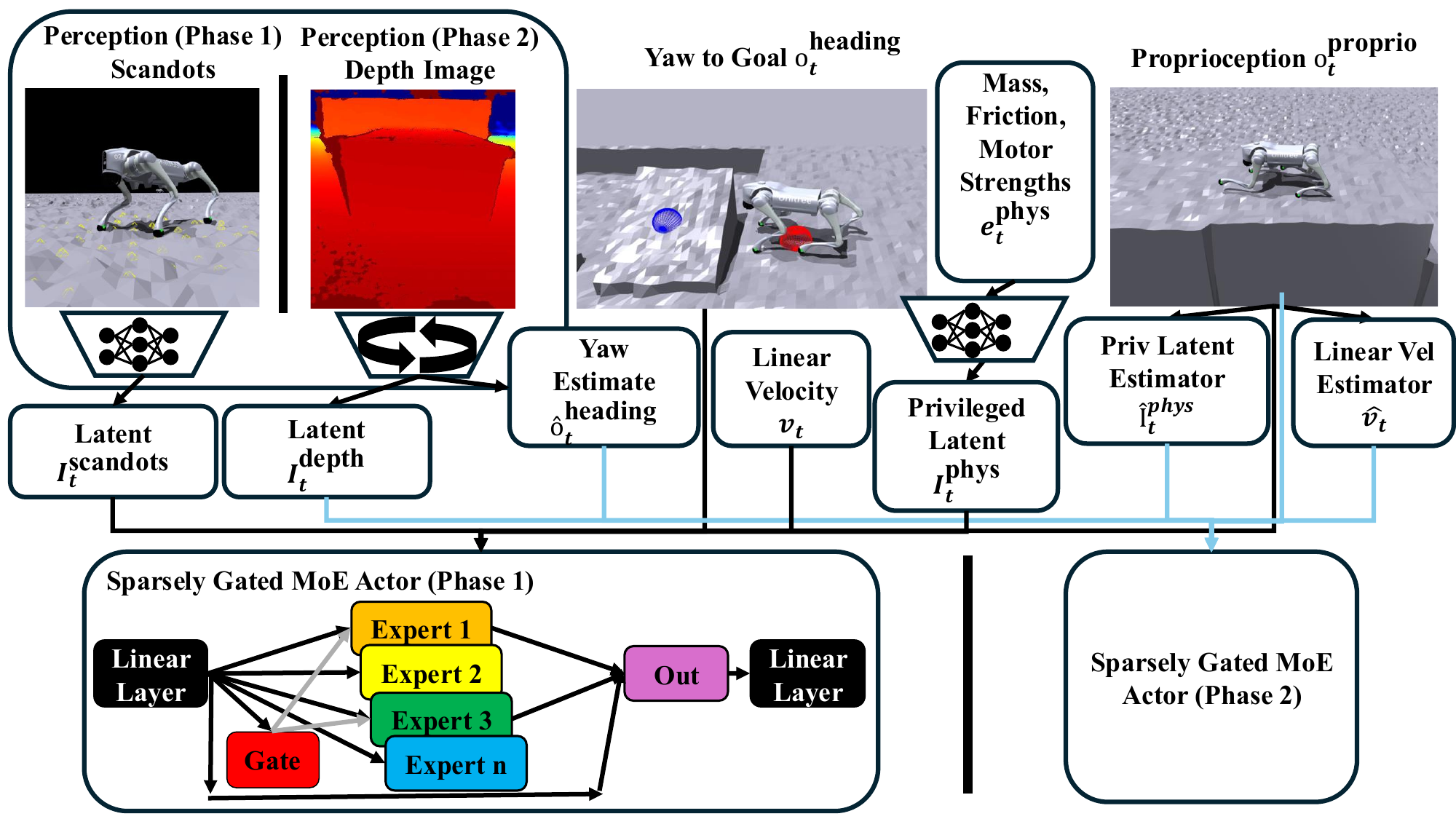}
    \caption{Model architecture during phase 1 and phase 2 training. Perception input is encoded with a multi-layer perceptron (MLP) during phase 1 and a Gated Recurrent Unit (GRU) during phase 2. During phase 2 the depth encoder is also used to output yaw estimates. Privileged information like center of mass, friction coefficients, and motor strengths are encoded during phase 1 and estimated in phase 2 from the proprioceptive history. Linear velocity is estimated in phase 2. The actor is a mixture of experts (MoE) that has a linear layer connected to the MoE layer that blends various experts sparsely according to a gating matrix, combines this with the residual, and passes the output to a final linear layer.}
    \label{fig:net_arch}
\end{figure*}

\subsection{Learning Parkour via RL}

The objective is to learn a policy $\pi$ parametrized by $\theta$ to control the robot in vision-based agile locomotion. To solve the problem, we employ reinforcement learning. The policy is used to sample an \textbf{action} $a_t \sim \pi_\theta(a_t|o_t)$ where $a_t \in \mathbb{R}^{12}$ is the joint position target of each of the 12 joints of the robot, and $o_t$ is the observation to be discussed in Sec.~\ref{subsec:obs}. The policy is trained to maximize expected rewards, defined in Sec.~\ref{Sec:Rewards}. We train purely in simulation, getting state transitions and observations from Isaac Gym~\cite{makoviychuk2021isaac} and a modified version of legged gym~\cite{rudin2022learning} with the depth pipeline. The training environment includes a series of boxes to climb, gaps to jump, and flat terrain, parkour pads to leap through, and hurdles to climb, arranged in increasing difficulty~\cite{cheng2024extreme}. Robots start in easier levels and as they traverse a certain amount of the terrain, are promoted to more difficult obstacles~\cite{agarwal2023legged}.

Our architecture is visualized in Fig.~\ref{fig:net_arch}. Much attention is paid to observation and dealing with privileged information using two phase learning in Sec.~\ref{subsec:obs}. This observation flows to the MoE actor discussed in Sec.~\ref{subsec:moe}, which is used in an actor-critic algorithm, specifically proximal policy optimization (PPO)~\cite{schulman2017proximal} in phase 1. PPO's loss is augmented with a term to encourage diversity of experts discussed in section \ref{subsec:moe}. Backpropagation is used to train the actor linear layers, various parameters of the actor's MoE layer including the gating network, the noise matrix, and the parameters of the expert component networks, parameters of the critic network, parameters of the scan-encoder used to encode elevation maps in phase 1, and parameters of the privileged information encoder used to encode in phase 1. In the second phase the phase 1 actor is taken as the teacher and mean-squared-error loss on actions is used to train the student which is initialized from the teacher parameters, and a term for yaw loss from~\cite{cheng2024extreme} is included.

\subsection{Observation and Two-Phase Training}\label{subsec:obs}

We use two-phase training~\cite{miki2022learning,chen2020learning} to take advantage of privileged information to make the problem more tractable and speed 
up the learning. The \textbf{observation} therefore changes between phase 1 and phase 2 but in both phases $o_t$ is a concatenation of 5 components based on proprioception $o_t^{\text{proprio}}$, perception $o_t^{\text{perception}}$, heading $o_t^{\text{heading}}$, privileged physical information $e_t^{\text{phys}}$, and privileged robot information $e_t^{\text{robot}}$. Generally in phase 2 we replace privileged information which we have at training time (given the simulator has the ground truth) but would not have at deployment on the robot. Proprioception is the only component that remains unchanged from phase to phase. The proprioceptive input includes base angular velocity $\omega_t \in \mathbb{R}^3$, base orientation that includes roll and pitch $\phi_t \in \mathbb{R}^2$, forward velocity command $v_t^{\text{x\_target}} \in \mathbb{R}^1$,  joint positions $q_t \in \mathbb{R}^{12}$, joint velocities $\dot{q}_t \in \mathbb{R}^{12}$, past actions $a_{t-1} \in \mathbb{R}^{12}$, and boolean contacts for feet $c_t \in \mathbb{R}^4$. Further, to the observation for the actor of angular velocity, pitch, roll, joint positions, joint velocities, we add Gaussian noise. To the boolean foot contact observations we flip the observation with a given probability. The critic (with hidden dimensions of 512, 256, 128) receives observations free of noise. Noise parameters used are given in table \ref{tab:domain_rand}. The final component of proprioception in the observation is a rolling history of length 10 of past proproprioceptive observations $o_{t-1:t-10}^{\text{proprio}}$.

The perceptive component of the observation $o_t^{\text{perception}}$ changes from phase to phase because depth images take time to render and are not the most stable to learn from \cite{zhuang2023robot}. In particular we use privileged information of scandots (a heightmap around the robot) \cite{agarwal2023legged} encoded by an MLP (with hidden dimensions of 128, 64, 32) to produce a latent $I_t^{\text{scandots}} \in \mathbb{R}^{32}$. The MLP is optimized at the same time as the actor in the first phase. In the second phase this is replaced by the depth image encoded by a Gated Recurrent Unit (GRU) (with hidden size 512)~\cite{chung2014empirical}. Depth image processing is elaborated in Sec.~\ref{sec:depth_pipeline}.

The heading component of the observation $o_t^{\text{heading}} \in \mathbb{R}^2$ is based on the Unified Reward for Extreme Parkour~\cite{cheng2024extreme} that lets the robot pick the heading by using oracle yaw inputs to the current goal $\psi_t^{\text{goal}}$ and next goal $\psi_t^{\text{next\_goal}}$ in phase 1 and replacing these gradually with estimated yaw targets $\hat{o_t}^{\text{heading}}$ from the depth encoder in phase 2. The privileged physical information $e_t^{\text{phys}}$ is based on online adaptation~\cite{kumar2021rma} whereby we take privileged information of the center of mass offsets $e_t^{\text{COM}} \in \mathbb{R}^3$, and mass offset $e_t^{\text{mass}} \in \mathbb{R}^1$, friction coefficient $e_t^{\text{friction}} \in \mathbb{R}^1$, and motor strengths that are used to change the proportional and derivative terms in the simulated PD controller that takes joint angle goals and outputs torque $e_t^{\text{motor}} \in \mathbb{R}^{24}$. The true physical characteristics are randomized within ranges for each environment, and the ranges are given in Table~\ref{tab:domain_rand}. These are encoded in a latent vector by an MLP (with hidden dimensions of 64, 20) $I_t^{\text{phys}} \in \mathbb{R}^{20}$ that is optimized at the same time as the actor in phase 1 that enables the latent to vary at the start of training and discourages variation after 3000 steps. In phase 2 this is replaced by an estimate from an MLP $\hat{I}_t^{\text{phys}}$ trained using supervised learning to mimic this latent that takes proprioceptive history as the input. The last observation component is the privileged robot information $e_t^{\text{robot}}$ whereby we take linear velocity of the robot in phase 1 $v_t \in \mathbb{R}^3$ and replace this with an estimate $\hat{v_t}$ in phase 2 from an MLP trained using supervised learning to estimate linear velocity given proprioception $o_t^{\text{proprio}}$. In total, $o_t \in \mathbb{R}^{563}$.

\subsection{Mixture of Experts}\label{subsec:moe}

We use an actor network of 3 sequential layers where the middle layer is the MoE and other two layers are linear layers and ELU activation~\cite{clevert2015fast} is used between layers. We use layer sizes of 512, 256, 256. The output of the MoE layer is a weighted combination of the $n$ expert outputs $E_i(x)$ based on layer input $x$ and the gating value $G(x)_i$ \eqref{eq:MoE}. The experts are simple weight matrices similar to that proposed by Cho et al.~\cite{cho2014exponentially}. 
\begin{gather}
  y = \sum_{i=1}^n G(x)_i E_i(x) \label{eq:MoE}
\end{gather}
The MoE layer is a sparsely-gated mixture of the top $k$ of $n$ experts based on work by Shazeer et al.~\cite{shazeer2017outrageously} and we therefore take advantage of the conditional computation afforded by sparse gating. We review this work here to make this paper self-contained. The intermediate gate output $H(x)_i$ takes as the basis the product of the layer input $x$ and a trainable weight matrix $W_g$ and adds tunable noise to help balance expert utilization. $W_{\text{noise}}$ and $W_g$ are initialized to zero at the start of learning. Given we use sparisty we employ residual connections to the prior layer to encourage gradient flow whereby we add the output of the prior layer to the output of the MoE layer.
\begin{gather}
  H(x)_i = (x \cdot W_g)_i + \mathcal{N}(0, 1) \cdot \text{softmax}((x \cdot W_{noise})_i) \label{eq:h}
\end{gather}
Top-k gating is used \eqref{eq:topn} whereby gate values not in the top-k are set to $-\infty$.
\begin{gather}
  \text{KeepTopK}(v,k)_i                = \begin{cases}
    v_i \text{ if $v_i$ $\in$ top k elements of v} \\
    -\infty \text{ otherwise}
  \end{cases} \label{eq:topn}
\end{gather}
and therefore final gate values $G(x)_i$ are zero post the softmax which is responsible for creating sparsity \eqref{eq:g}.
\begin{gather}
G(x)_i = \text{softmax}(\text{KeepTopK}(H(x)_i, k)) \label{eq:g}
\end{gather}
To encourage diversity of experts given MoEs can favor a limited number of experts a load balancing loss term, again from Shazeer et al.~\cite{shazeer2017outrageously} is adopted into phase 1 and phase 2 losses where the importance of an expert is calculated as the sum of its weights across the batch \eqref{eq:imp}
\begin{gather}
\text{Importance}(X) =\sum_{x \in X}  G(x) \label{eq:imp}
\end{gather}
and the coefficient of variation is calculated as the standard deviation of importances divided by the mean of importances and this is weighted with a manually tuned coefficient $w_{\text{Importance}}$ to derive the additional loss term \eqref{eq:loss_imp}. We use a $w_{\text{Importance}}$ of 0.05, $k$ of 4, and $n$ of 16.
\begin{gather}
    L_{\text{Importance}}(X) = w_{\text{Importance}} \cdot \text{CV}(\text{Importance}(X))^2 \label{eq:loss_imp}
\end{gather}

\begin{table}[]
    \centering
    \caption{Domain randomization. Distribution type, u for uniform between bounds low $l$ and high $h$, g for gaussian parametrized by a mean $\mu$ and sigma $\sigma$, 
    and b for a binomial distribution parametrized by a probability $p$.}
    \renewcommand\arraystretch{1.1}
    \renewcommand\tabcolsep{4.0pt}
    \footnotesize
    \begin{tabular}{c|c|c|c|c|c|c}
        \toprule[0.03cm]
        \tb{Term}   & \tb{Type}            & \tb{$l$} & \tb{$h$} & \tb{$\mu$} & \tb{$\sigma$} & \tb{$p$}
        \\
        \midrule[0.03cm]
        rotation           & g & - & -& 0& 0.025& -\\
        joint\_pos           & g & - & -& 0& 0.01& -\\
        joint\_vel           & g & - & -& 0& 1.5& -\\
        ang\_vel           & g & - & -& 0& 0.2& -\\
        foot\_contact           & b & - & -& -& -& 0.05\\
        cam\_x\_pos           & g & - & -& 0.32& 0.001& -\\
        cam\_y\_pos           & g & - & -& 0.017& 0.001& -\\
        cam\_z\_pos           & g & - & -& 0.15& 0.001& -\\
        cam\_x\_rot           & u & -1& 1& -& -& -\\
        cam\_y\_rot           & u & 22.5& 24& -& -& -\\
        cam\_z\_rot           & u & -0.25& 0.25& -& -& -\\
        horizontal\_fov          & u & 86.5& 87.5& -& -& -\\
        depth\_artifact          & b & -& -& -& -& 0.001\\
        depth\_artifact\_height          & g & -& -& 3& 3& -\\
        depth\_artifact\_width          & g & -& -& 3& 3& -\\
        contour\_artifact          & b & -& -& -& -& 0.1\\
        gaussian\_blur\_sigma          & u & 1.5& 1.5& -& -& -\\
        $v_t^{\text{x\_target}}$          & u & 0.3& 0.8& -& -& -\\
        $e_t^{\text{friction}}$          & u & 0.6& 2.0& -& -& -\\
        $e_t^{\text{COM}}$          & u & -0.2& 0.2& -& -& -\\
        $e_t^{\text{mass}}$          & u & 0.0& 3.0& -& -& -\\
        $e_t^{\text{motor}}$          & u & 0.8& 1.2& -& -& -\\
        \bottomrule[0.03cm]
    \end{tabular}
    \label{tab:domain_rand}
    \vspace{-0.35cm}
\end{table}

\subsection{Rewards}\label{Sec:Rewards}
\textbf{Rewards} are given in Table~\ref{tab:exp_rpe_map_size}. Variable definitions are provided: $\boldsymbol{r}_t$ is the ground truth orientation of the robot in quaternion form, the projected gravity quaternion is $\boldsymbol{g}_t$, the timestep $\delta t$ is 50HZ, $f_t^j$ indicates the jth element of the robot model and its contact force, a subset of robot elements including the base, head, thigh, and calf is used for collision indices, the torque on the ith motor is $\tau_t^i$, the feet\_edge reward is taken from \cite{cheng2024extreme} using $M$ a boolean function indicating if a foot position $p_i^{\text{foot}}$ is near an edge in the environment, the tracking\_goal\_vel is also from \cite{cheng2024extreme} and uses $\hat{d}_w=\|p^{\text{robot}} - p^{\text{goal}} \|_2$ where $p$ is the position in world coordinates, the default position for each joint is $w_t$. In training, to avoid early termination problems, we only use positive rewards by flooring rewards for a trajectory at zero.


\begin{table}[]
    \centering
    \caption{Rewards terms used including coefficients and their applications.}
    \renewcommand\arraystretch{1.1}
    \renewcommand\tabcolsep{4.0pt}
    \footnotesize
    \begin{tabular}{c|c|c}
        \toprule[0.03cm]
        \tb{Term}         & \tb{Coefficient} & \tb{Formula}
        \\
        \midrule[0.03cm]
        tracking\_yaw& 0.5 & $e^{-|\psi_t^{\text{goal}} - \psi_t|}$\\
        lin\_vel\_z & -1.5 & $\begin{cases}
    v_t^z \text{ if walking env} \\
    0.5 v_t^z \text{ otherwise}
  \end{cases}$\\
        ang\_vel\_xy  & -0.05 & $(\omega_t^x)^2 + (\omega_t^y)^2 $\\
        orientation      & -1.0 & $\begin{cases}
    (g_t^x)^2 + (g_t^y)^2 \text{ if walking env} \\
    0 \text{ otherwise}
  \end{cases}$\\
        dof\_acc      & -2.5e-7 & $\sum_{i=0}^{12}((\dot{q}_t^i - \dot{q}^i_{t-1} )/ \delta t)^2$\\
        collision     & -10. & $ \sum_{j \in \text{collision\_bodies}} \begin{cases}
    1.0 \text{ if } |f_t^j|_2 > 0.1 \\
    0 \text{ otherwise}
  \end{cases}$\\
        action\_rate      & -0.1 & $\|a_t - a_{t-1}\|_2$\\
        delta\_torques        & -1.0e-7 & $\sum_{i=0}^{12} (\tau_t^i - \tau_{t-1}^i)^2$\\
        torques        & -1e-5 & $\sum_{i=0}^{12} (\tau_t^i)^2$\\
        hip\_pos         & -0.5 & $\sum_{i \in \text{hip index}} (q_t^i - w_t^i)^2$\\
        dof\_error        & -0.04 & $\sum_{i} (q_t^i - w_t^i)^2$\\
        feet\_stumble    & -1.0 & $\begin{cases}
    1 \text{ if }  \Sigma_{j \in \text{feet}} \|f_j^{\text{x,y}}\|_2 > 4 * |f_j^z| \\
    0 \text{ otherwise}
  \end{cases}$ \\
        feet\_edge      & -1.0 & $\sum_i c_i \cdot M[p_i]$\\
        tracking\_goal\_vel & 1.5 & $\text{argmin}(\langle v_t, \hat{d}_w\rangle, v_t^{\text{x\_target}}) )$\\
        \bottomrule[0.03cm]
    \end{tabular}
    \label{tab:exp_rpe_map_size}
    \vspace{-0.35cm}
\end{table}

\subsection{Depth Image Pipeline} \label{sec:depth_pipeline}

\begin{figure*}
    \centering
    \begin{subfigure}{1.0\textwidth}
        \begin{subfigure}{0.195\textwidth}
        \includegraphics[width=1.0\textwidth]{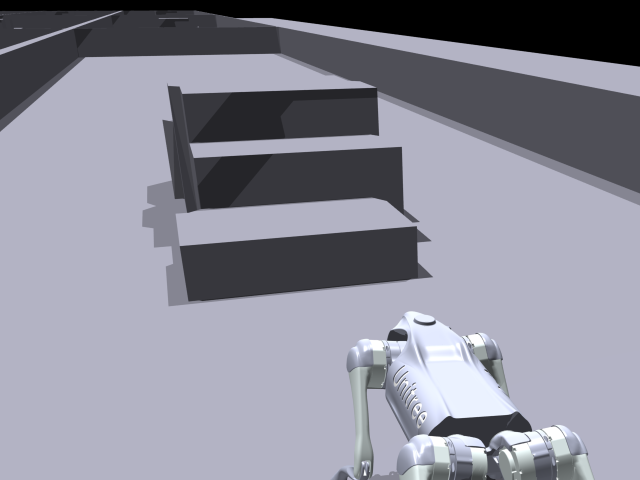}
        \caption{A similar scene.}
        \label{fig:depth_real}
        \end{subfigure}
        \begin{subfigure}{0.195\textwidth}
        \includegraphics[width=1.0\textwidth]{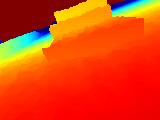}
        \caption{Rendered depth.}
        \label{fig:depth_real}
        \end{subfigure}
        \begin{subfigure}{0.195\textwidth}
        \includegraphics[width=1.0\textwidth]{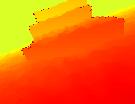}
        \caption{Clip, contour, crop.}
        \label{fig:depth_real}
        \end{subfigure}
        \begin{subfigure}{0.195\textwidth}
        \includegraphics[width=1.0\textwidth]{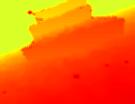}
        \caption{Artifacts, blur.}
        \label{fig:depth_real}
        \end{subfigure}
        \begin{subfigure}{0.195\textwidth}
            \begin{subfigure}{0.45\textwidth}
            \includegraphics[width=1.0\textwidth]{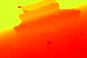}
            \caption{Resize image.}
            \label{fig:depth_real}
            \end{subfigure}
            \begin{subfigure}{0.45\textwidth}
            \includegraphics[width=1.0\textwidth]{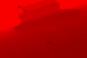}
            \caption{Normalize.}
            \label{fig:depth_real}
            \end{subfigure}
        \end{subfigure}
        
    \end{subfigure}
    
    \caption{Depth images that are rendered start off with perfectly correct information. We degrade it in various steps to add noise that is similar to the noise seen on the real hardware, an Intel Realsense D435f Depth Camera.}
  \label{fig:depth_processing_pipeline}
\end{figure*}

Another sim-to-real gap exists in the depth images rendered in simulation and the depth images available from the robot. To bridge 
this gap when training in phase 2 we attach the depth cameras to the robots while varying the position, angle, and field of view of the cameras within the bands given in Table~\ref{tab:domain_rand}. While stepping the environment and collecting 
sensor data we also pre-process the images. In particular we follow the steps visualized in Fig.~\ref{fig:depth_processing_pipeline} where when deploying on the robot we skip depth contour and 
artifact steps. 

We take in images with resolution (160, 120). First, we clip images to the minimum (0.15 meters) and maximum (3 meters). Second, we find contours of the image with a threshold (1.0) on the depth gradient and set found contours to the minimum (0.15 meters) with a probability given in \ref{tab:domain_rand}. Third, we crop the image taking 20 pixels from the left, 5 from the right, 16 from the bottom, and none from the top. Fourth, we add artifacts to the image with a given probability with artifact width and height parametrized by Gaussians, probability and parameters given in Table~\ref{tab:domain_rand}. Fifth, we employ Gaussian blurring with a kernel of size 3 and a sigma parametrized in Table~\ref{tab:domain_rand}. Sixth, we downsample the image to the size (87, 58). Seventh, we normalize the image according to the minimum and maximum clipping and centering around 0.

\section{Results}\label{Sec:Results} 

We compare the MoE to baselines built on Extreme Parkour but with our image pipeline. The baselines use a sequential MLP with ELU activation between layers for the actor and vary the number of parameters to test the hypothesis that MoE helps scale performance. We test training time performance in \ref{Sec:small}, performance on the real Unitree Go2 Robot in \ref{Sec:real_res}, and we investigate expert specialization in \ref{Sec:expert_spec}.



\subsection{Hardware}

We use a Unitree Go2 Robot with an Intel Realsense D435f depth camera mounted. This camera includes an IR filter that helped filter 
out the reflections of lights on the floor of the laboratory seen in Fig.~\ref{fig:parkour_composite}. We also designed and fabricated a camera mount 
for the Go2 robot that is available in the public code repository. The camera's angle of orientation around the y-axis helps
determines its view and 23 degrees was chosen. It is also important for the mount to have rigid attachment to the robot to avoid camera movement 
relative to the robot during dynamic motion.  We publish depth images at 15HZ from the robot's onboard computer but 
run the policy off-board on an Nvidia RTX 2060 GPU. When deploying on the real robot we clip actions given torque limits, position goals, and current positions, mimicking the PD controller running on the robot~\cite{zhuang2023robot}. We also kill the program if pitch, roll, joint position, or joint torque are out of pre-set boundaries. We train on an Nvidia RTX 4090 GPU. We use 40 for the proportional gain and 1 for the derivative gain in the motor PD controllers.

\subsection{Training Metrics}\label{Sec:small}

We train policies with varying actor architectures, measure training time metrics for phase 1, and summarize results in table \ref{tab:micro_results}. The policy that trains one policy per skill (Policy/Skill) in a traditional approach to distilling new skills from experts does not perform well with the lowest reward and terrain level of any policy. This may be due to the fixed number of environments of 6,000, where each skill expert didn't see enough examples to specialize and train an entire actor network. The sequential MLP policies that match inference time parameters and total parameters under-perform the MoE with respect to reward and terrain level, though the larger MLP does get close to the MoE. Intuitively, adding parameters to a network when the problem is tough and learning may benefit from more capacity should help performance. The MoE is able to add parameters but thanks to sparsity does this without increasing inference time parameters an equivalent amount which may contribute to performance. There are also expert specialization effects in latent space that may also contribute \ref{Sec:expert_spec}.

\begin{table}
    \centering
    \caption{Training time metrics are summarized for various policies including: an MoE architecture; a sequential MLP architecture that matches inference time parameters of the MoE; a sequential MLP that matches total parameters of the MoE; and a policy that trains one Sequential MLP for each skill. Average reward during the last 1000 timesteps of phase 1 training is presented with average terrain level during the same timesteps. The MoE shows out-performance with respect to terrain level and reward..}
    \renewcommand\arraystretch{1.1}
    \renewcommand\tabcolsep{4.0pt}
    \footnotesize
    \begin{tabular}{c|c|c|c|c}
        \toprule[0.03cm]
        \tb{Algorithm}  & \tb{Param} & \tb{Inf. Param} & \tb{Reward} & \tb{Terrain Lvl} 
        \\
        \midrule[0.03cm]
        \tb{MoE top 4/16} & $153 \times 10^3$ & $102 \times 10^3$& \tb{10.29}& \tb{5.73} \\ 
        Sequential & $102 \times 10^3$ & $102 \times 10^3$& 9.51& 5.53 \\ 
        Sequential & $153 \times 10^3$ & $153 \times 10^3$& 9.91& 5.62 \\ 
        Policy/Skill& $500 \times 10^3$& $500 \times 10^3$& 8.17 & 1.32\\ 
        \bottomrule[0.03cm]
    \end{tabular}
    \label{tab:micro_results}
    \vspace{-0.35cm}
\end{table}

\subsection{Real Robot Performance Tests}\label{Sec:real_res}

We also train an MoE policy and sequential MLPs to deploy on the Unitree Go2. The MoE has hidden layers of sizes 512, 256, 256. The varying hidden layer sizes of the sequential MLP actors are: 512, 512, 387 which is roughly the number of parameters the MoE has at inference time; 1024, 620, 512 which is roughly the number of total parameters the MoE has. We test on a 15cm Box that is an easier obstacle and a 32cm box that is 80\% of the robot's height seen in Fig.~\ref{fig:parkour_composite}, a challenging obstacle. We deploy and test each policy multiple times. For policies with low robustness we test 5 times for safety reasons, for other policies we test 10 times. The number of trials is recorded as is the number of successful trials where the robot overcomes the obstacle. Table~\ref{tab:main_results} 
summarizes the results. Fig.~\ref{fig:parkour_composite} shows one of the 5 runs for the MoE policy with depth and color images from the camera where the robot approaches the obstacle, hops on-top, and then dismounts.  

The MoE and the sequential policy matching inference time parameters perform similarly in the real-world tests. The performance difference may be too small and the tests too coarse to bring out the differences in the policies. Interestingly the sequential policy matching total parameters does not perform well which could indicate overfitting.

\begin{table*}
    \centering
    \caption{Sequential MLP policies with the same number of parameters as the MoE at inference time have similar performance on the obstacles with the real robot. Sequential policies that match total parameters have worse performance than the MoE which may indicate overfitting.}
    \renewcommand\arraystretch{1.1}
    \renewcommand\tabcolsep{4.0pt}
    \footnotesize
    \begin{tabular}{c|c|c|c|c|c|c}
        \toprule[0.03cm]
        \tb{Algorithm}  & \tb{Param} & \tb{Inf. Param} & \tb{\# Trials 15cm Box} & \tb{\# Success 15cm Box} & \tb{\# Trials 32cm Box} & \tb{\# Success 32cm Box}
        \\
        \midrule[0.03cm]
        \tb{MoE top4/16}& $1.5\times10^{6}$& $0.8\times10^{6}$& 10& 10& 10& 6 \\
        Sequential & $0.8\times10^{6}$ & $0.8\times10^{6}$ & 10& 10& 10 & 6 \\ 
        Sequential  & $1.5\times10^{6}$ & $1.5\times10^{6}$ & 10& 9& 5& 0 \\ 
    \end{tabular}
    \label{tab:main_results}
    \vspace{-0.35cm}
\end{table*}

\subsection{Expert Specialization}\label{Sec:expert_spec}

\begin{figure*}[t]
    \begin{subfigure}{0.49\textwidth}
        \includegraphics[width=1.0\textwidth, trim={5 5 10 10},clip]{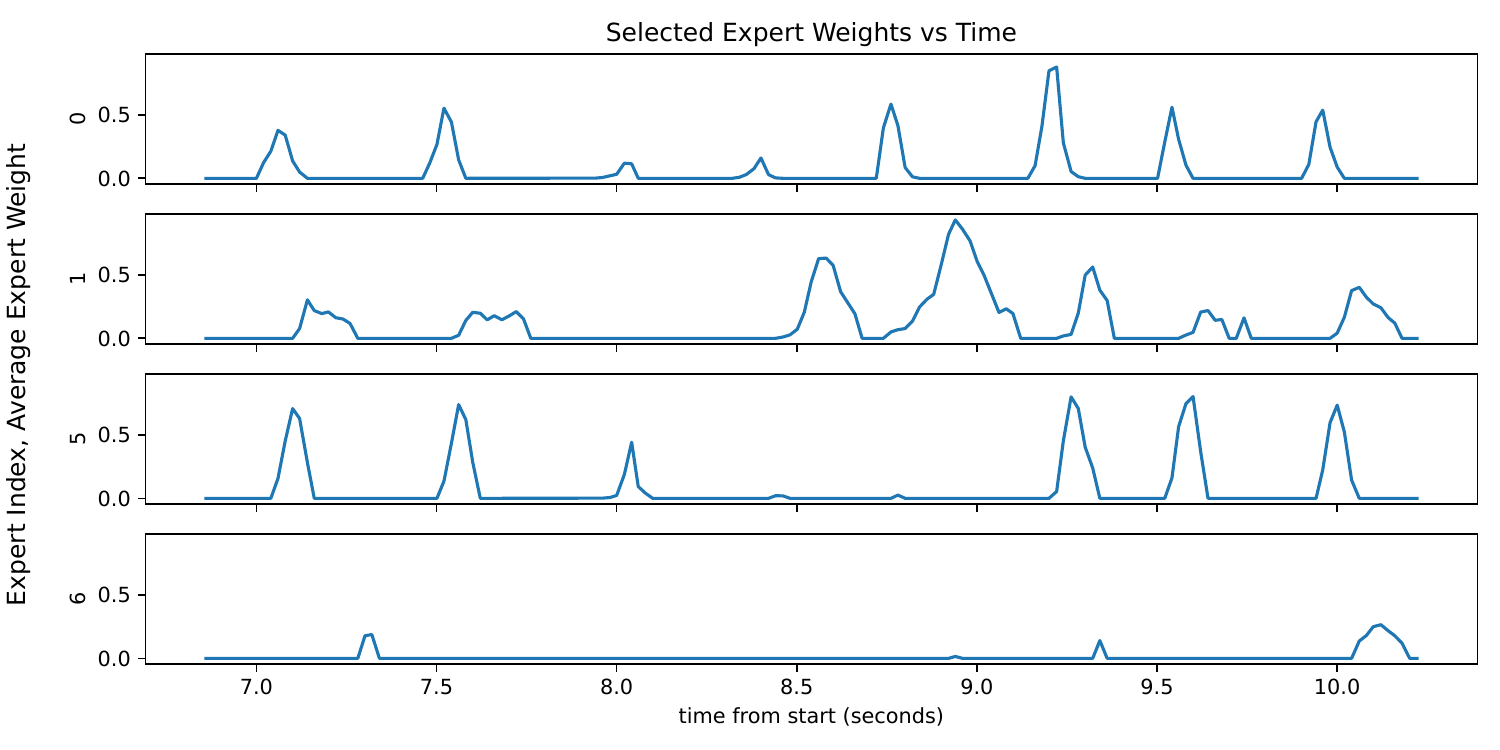}
        \caption{}
        \label{fig:weight_real}
    \end{subfigure}
    \begin{subfigure}{0.49\textwidth}
        \includegraphics[width=1.0\textwidth, trim={5 5 10 10},clip]{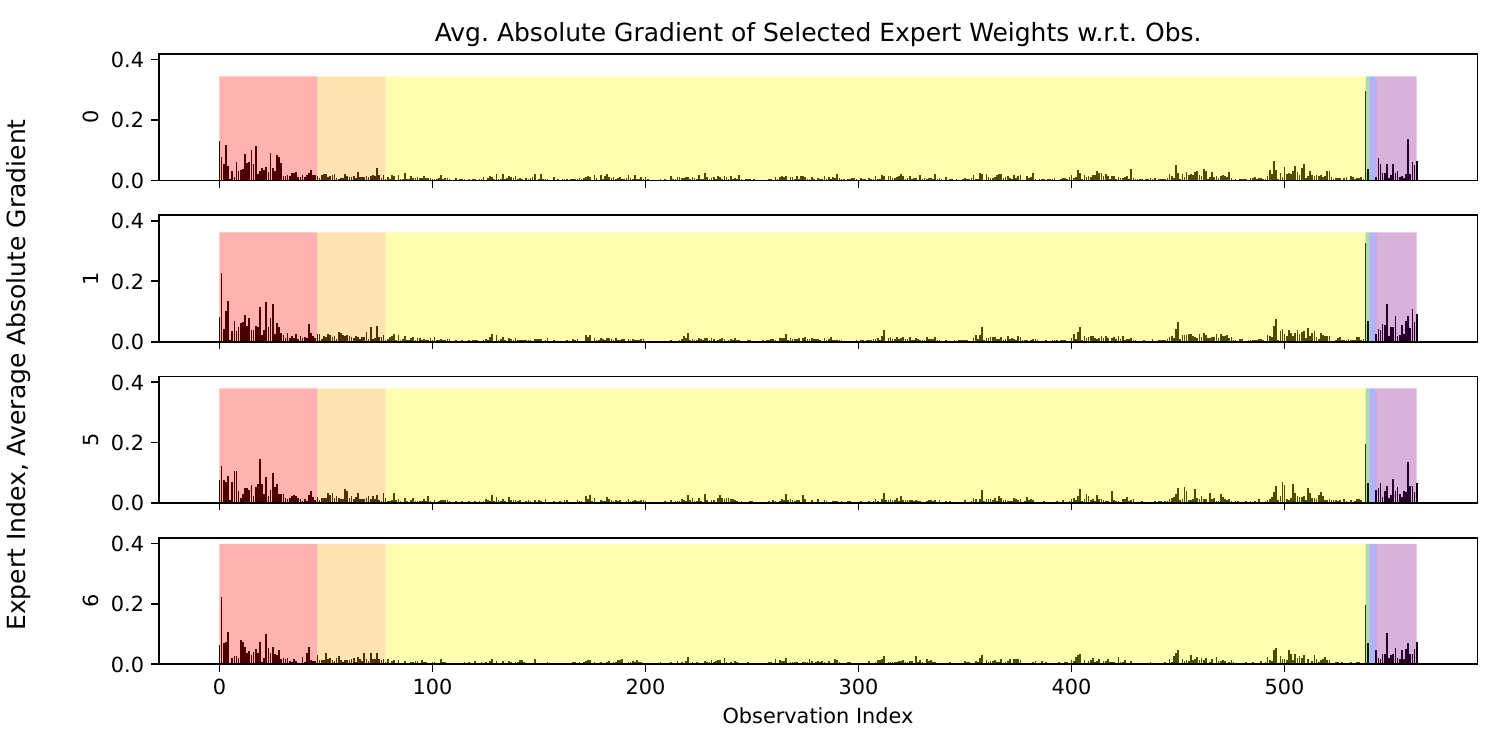}
        \caption{}
        \label{fig:weight_grad}
    \end{subfigure}
    \caption{Selected expert weights throughout the trial in figure \ref{fig:parkour_composite} are analyzed. Figure \ref{fig:weight_real} shows strong cyclical signals for most experts which is expected given the cyclical nature of locomotion and indicates expert specialization at different points in the cycle. Expert index 6 is the least used across experts with non-cyclical weighting. Figure \ref{fig:weight_grad} analyzes the gradient of weights whereby proprioception $o_t^{\text{proprio}}$ is red, depth latent is orange $I_t^{\text{depth}}$, proprioceptive history is yellow $o_{t-1:t-10}^{\text{proprio}}$, yaw is green $\hat{o}_t^{\text{heading}}$, robot latent is blue $\hat{I}_t^{\text{robot}}$, physics latent is purple $\hat{I}_t^{\text{phys}}$. Notably while expert index 5 is the most sensitive to the depth latent, many other experts are also sensitive. Also of note is that more recent proprioceptive observations in the buffer are favored more heavily (higher index is more recent). Yaw is a big driver of expert weighting.}
  \label{fig:expert_weight_analysis}
\end{figure*}

We also analyze expert weighting. Across the trial in Fig.~\ref{fig:parkour_composite} experts have utilization from 6.5\% to 41.4\% of timesteps where expert index 6 is the least used and expert index 1 is the most used. The selected experts in Fig.~\ref{fig:weight_real} show strong cyclical signals in expert weights in a trial with the exception of expert index 6. Locomotion is cyclical, so this may indicate expert specialization in aspects of the locomotion cycle. Analysis in Fig.~\ref{fig:weight_grad} analyzes the gradient of expert weights by playing back the recorded data from the trial in Fig.~\ref{fig:parkour_composite} and running inference on observations tracking expert weights and the gradient with respect to the observation. For each expert non-zero gradients are collected across timesteps and the average absolute gradient for each component of the observation is taken across timesteps. Notably there is significant attention placed on yaw, the depth latent, the observation history.  In contrast to architectures that train one policy per skill and then switch between policies depending on the skill the MoE architecture here specializes implicitly in latent space and many experts are used in climbing the obstacle.

\subsection{Outdoor Tests}

To test the performance of the MoE policy outside laboratory conditions we deploy the robot in a skate-park with obstacles that Fig.~\ref{fig:parkour_demos} shows. The robot is able to overcome such obstacles outside of laboratory conditions with glare from sunlight and cluttered ground.

\section{Conclusion}\label{Sec:Concl}

We present a mixture of experts system for vision-based locomotion leveraging conditional compute to scale performance. We analyze training metrics and find benefits in the proposed architecture. We also deploy the MoE onto a Unitree Go2 Robot and find real-world performance. However, the MoE system introduces many more hyper-parameters that need to be tuned like the number of experts, the number of top experts to filter for conditional 
computation, and the coefficient on the loss to encourage expert utilization. Future work may build on this by applying MoE to tasks with a need for network capacity or apply other recent techniques from LLMs to locomotion.

\section{Acknowledgments}\label{Sec:Ack}
Thanks to Chelsea Finn for her guidance in reinforcement learning and research. Thanks to Danny Steadman for fabrication support on the camera mount, to Simon Gosling for help on box design, and to John Osmotherly and Massimo Aurelio Cappella for training on woodworking tools and help with that big machine in the back.

\bibliographystyle{IEEEtran}
\bibliography{references}

\end{document}